\documentclass{article}
\usepackage{fancyhdr}

\usepackage{report}

\usepackage{soul}
\usepackage[utf8]{inputenc} 
\usepackage[T1]{fontenc}    
\usepackage{hyperref}       
\usepackage{url}            
\usepackage{booktabs}       
\usepackage{amsfonts}       
\usepackage{fancyhdr}       

\usepackage{nicefrac}       
\usepackage{microtype}      
\usepackage{graphicx}
\usepackage{pifont}
\usepackage{multirow}
\usepackage{CJKutf8}
\definecolor{darkmagenta}{rgb}{0.56, 0.0, 1.0}
\definecolor{softyellow}{rgb}{1.0, 0.92, 0.3} 
\definecolor{LightAquamarine}{rgb}{0.75, 1.0, 0.8} 
\definecolor{FireBrick}{RGB}{178,34,34}
\definecolor{MediumPurple}{RGB}{147,112,219}

\definecolor{uclablue}{rgb}{0.15, 0.45, 0.68}
\hypersetup{
    breaklinks,
    colorlinks=true,
    citecolor={darkmagenta},
    linkcolor={uclablue},
    urlcolor={uclablue}
}
\usepackage{wrapfig}
\usepackage{float}
\usepackage{subcaption}
\usepackage{placeins}
\usepackage{lipsum} 
\usepackage{tcolorbox}
\usepackage{amsmath}
\usepackage{amssymb}
\usepackage{utfsym}
\usepackage{fontawesome}
\usepackage{xspace}
\usepackage{enumitem}
\usepackage{multirow} 
\tcbuselibrary{breakable}
\usepackage{enumitem}
\usepackage{colortbl}
\usepackage{fancyhdr}

\usepackage{tcolorbox}
\usepackage{transparent}

\definecolor{njuPurple}{RGB}{220,205,230}     
\definecolor{njuPurpleLight}{RGB}{250,245,252}   

\newtcolorbox{abstractbox}{
    colback=njuPurpleLight,   
    colframe=njuPurple,       
    boxrule=1pt,              
    arc=4mm,                  
    left=8pt,                 
    right=8pt,                
    top=8pt,                  
    bottom=8pt,               
    opacityback=0.95
}

\title{Vibe AIGC: A New Paradigm for Content Generation via Agentic Orchestration}

\author{
\textbf{Jiaheng Liu$^{1}$},
\textbf{Yuanxing Zhang$^{2}$},
\textbf{Shihao Li$^{1}$},
\textbf{Xinping Lei$^{1}$} \\
\vspace{4mm}
{\normalsize $^1$ NJU-LINK Team, Nanjing University} \quad
{\normalsize $^2$ Kling Team, Kuaishou Technology} \\ 
\vspace{2mm}
\texttt{liujiaheng@nju.edu.cn} \\
}

\begin{document}

\maketitle

\begin{abstractbox}
\begin{center}
\textbf{\Large Abstract}
\end{center}
For the past decade, the trajectory of generative artificial intelligence (AI) has been dominated by a model-centric paradigm driven by scaling laws. Despite significant leaps in visual fidelity, this approach has encountered a ``usability ceiling'' manifested as the Intent-Execution Gap (i.e., the fundamental disparity between a creator's high-level intent and the stochastic, black-box nature of current single-shot models). In this paper, 
inspired by the Vibe Coding, we introduce the \textbf{Vibe AIGC}, a new paradigm for content generation via agentic orchestration, which represents the autonomous synthesis of hierarchical multi-agent workflows. 
Under this paradigm, the user's role transcends traditional prompt engineering, evolving into a Commander who provides a Vibe, a high-level representation encompassing aesthetic preferences, functional logic, and etc. A centralized Meta-Planner then functions as a system architect, deconstructing this ``Vibe'' into executable, verifiable, and adaptive agentic pipelines. By transitioning from stochastic inference to logical orchestration, Vibe AIGC bridges the gap between human imagination and machine execution. We contend that this shift will redefine the human-AI collaborative economy, transforming AI from a fragile inference engine into a robust system-level engineering partner that democratizes the creation of complex, long-horizon digital assets.
\end{abstractbox}

\begin{figure*}[h]
    \centering
    \includegraphics[width=1\linewidth]{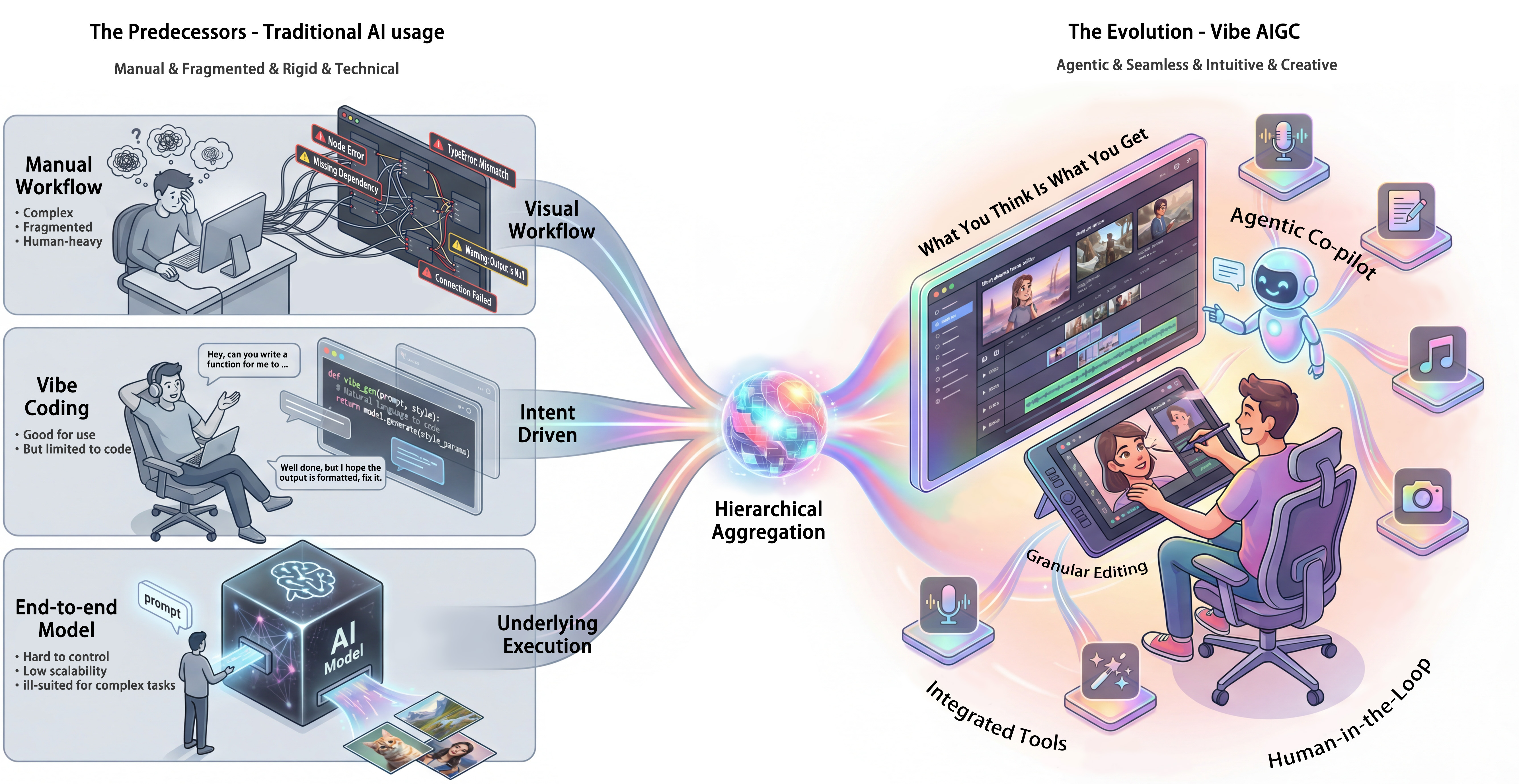}
    \caption{Content generation advancing toward the Vibe AIGC era: A systemic leap driven by structural combination.}
    \label{fig:intro}
\end{figure*}

\section{Introduction}

The trajectory of generative artificial intelligence (AI) has reached a critical juncture. For the past decade, the community has operated under a ``Model-Centric'' paradigm~\citep{blattmann2023stable, achiam2023gpt, rombach2022high}, where progress is primarily measured by the expansion of parameter counts, the ingestion of increasingly massive datasets, and the refinement of end-to-end training objectives. From the initial breakthroughs in Generative Adversarial Networks (GANs)~\citep{Goodfellow2014GenerativeAN} to the current dominance of Diffusion Transformers (DiTs)~\citep{peebles2023scalable}, the prevailing belief has been that ``scaling laws'' would eventually bridge the gap between human imagination and machine execution. However, as these foundational models are deployed into professional creative environments—ranging from cinematic production to complex narrative synthesis—a fundamental ``usability ceiling'' has emerged. Despite the undeniable increase in visual fidelity, the actual process of content creation remains a fragile exercise in stochastic trial-and-error.

The root of this ceiling lies in the \textbf{Intent-Execution Gap}: \textit{the inherent disparity between a human creator’s high-level, multi-dimensional vision and the ``black-box'' nature of current single-shot generation}. In today’s AIGC (i.e., Artificial Intelligence Generated Content) landscape, the user is relegated to the role of a ``prompt engineer''—a digital manual laborer who must spend hours performing ``latent space fishing'', hoping that a specific string of keywords will align with the model's internal weights to produce a coherent result. This workflow is fundamentally unscalable for professional applications that require temporal consistency, character fidelity, and deep semantic understanding. Even as models become larger, they remain architecturally flat; they lack the hierarchical reasoning and iterative verification loops necessary to manage long-horizon creative tasks. When a model hallucinates a detail—such as the incorrect school uniform in a commemorative video or a disjointed narrative arc—the user is often left with no recourse but to re-roll the generation, a process that is both computationally wasteful and creatively frustrating.

As shown in Figure~\ref{fig:intro},
we observe that the field of software engineering is currently undergoing a radical transformation known as ``Vibe Coding''~\citep{karpathy2025}, where natural language is being utilized not as a mere interface for code, but as a high-level kernel for autonomous system construction~\citep{mei2025survey}. We believe the generative AI community is on the cusp of a similar, yet even more profound, transition. It is no longer sufficient to treat content generation as a single-pass inference problem. Instead, we must begin to view the generation of complex media as a system-level engineering challenge that requires the synthesis of specialized agentic behaviors.

In this paper, we argue that the current research focus on end-to-end ``one-size-fits-all'' models is reaching a point of diminishing returns for the human-AI collaborative economy~\citep{bai2023qwentechnicalreport}. {We contend that the machine learning community must pivot its fundamental research objective: shifting away from Model-Centric Generation toward \textbf{Vibe AIGC}, a paradigm in which content generation is reconceptualized as the autonomous synthesis of multi-agent workflows driven by high-level human intent.}

Specifically,
in Vibe AIGC, we believe that the next frontier of artificial intelligence is not larger models, but smarter orchestration, and propose a transition where the user moves from ``Prompt Engineer'' to ``Commander'', providing the ``Vibe'' environment, a high-level representation of aesthetic, logic, and intent, where a Meta-Planner then deconstructs into executable and verifiable multi-agent pipelines~\citep{liu2023pre, horvat2025vibe}.

In the following paper, we first explore the philosophical foundations of Vibe Coding (Section~\ref{vibecoding}). We then provide a technical critique of current model-centric architectures (Section ~\ref{modelcentric}). Drawing on preliminary successes in agentic frameworks (Section~\ref{attempts}), we detail the top-level architecture of Vibe AIGC (Section~\ref{vibeaigc}), emphasizing the role of hierarchical orchestration. Finally, we introduce the alternative views (Section~\ref{views} and the call-to-action for building the Vibe AIGC ecosystem (Section~\ref{action}).

\section{Vibe Coding}
\label{vibecoding}

\subsection{Definition of Vibe Coding}
In the history of computer science, the evolution of programming has been a steady march away from machine hardware toward human cognition—from assembly to C, and from C to Python~\citep{gemini-code-cli,zeng2025glm,anthropic2025claude45,achiam2023gpt, liu-etal-2025-m2rc}. Each step increased the level of abstraction, allowing developers to ignore low-level complexities to focus on logic.
The term ``Vibe Coding'' popularized by researchers like Andrej Karpathy~\citep{karpathy2025, horvat2025vibe}, represents the logical conclusion of this trajectory: the removal of formal syntax entirely.
In this framework, the ``Vibe'' refers to a high-level, multi-dimensional representation of intent that includes aesthetic preference, functional goals, and systemic constraints. Unlike a traditional prompt, which is often a one-shot instruction~\citep{mei2025survey, sapkota2025vibe}, a ``Vibe'' is a continuous latent state maintained through dialogue. We argue that natural language has reached a critical threshold of semantic density where it can function as a ``meta-syntax''. Within a Vibe Coding environment, the AI does not just execute a command,
and it interprets the ``atmosphere'' of the project to make autonomous decisions, such as selecting appropriate library dependencies or adhering to an unstated but implied design language that previously required human intervention.

\subsection{User as Commander}
The most profound shift in Vibe Coding is the reconfiguration of the human user’s identity~\citep{Chen2025ScreenReader}. Throughout the Model-centric era for AIGC, the user was essentially a manual laborer of the interface (i.e., a prompt engineer, who spent hours in a stochastic trial-and-error loop, attempting to find the specific ``magical'' string of text that would yield a desired result). This role is inherently limited by the user’s ability to predict the model’s internal weights.
Vibe Coding proposes a transition where the user acts as a ``Commander'' (or Architect). In this paradigm, the human provides the strategic vision (\textit{the What and the Vibe}), while the AI system autonomously determines the tactical implementation (\textit{the How}). This is analogous to the shift from a pilot manually controlling every flap on an aircraft to a commander setting a destination on an advanced autopilot system. By delegating the low-level generation of code or assets to agentic workflows, the user can operate at the level of system design. 
{This democratization is crucial, as it allows individuals with domain expertise to command complex digital systems, thereby expanding the creative and economic potential of the digital economy}.

\subsection{Agentic Orchestration}
A primary failure of current AIGC tools is the ``Intent-Execution Gap'' (i.e., the disparity between a complex creative vision and the flattened, often mediocre output of a single-shot model).
Vibe Coding addresses this gap not through better base models, but through recursive orchestration~\citep{dang2025multi}.
In a Vibe-driven system, the AI does not attempt to solve a complex problem in one pass. Instead, it uses the ``Vibe'' as a compass to synthesize a custom, multi-step workflow. For instance, if a user wants to create a ``vibrant, cinematic music video'', a Vibe Coding agent does not simply call a video-generation API. It recursively breaks the vibe into constituent parts, and it writes a script, analyzes the musical tempo, generates character consistency sheets, and oversees the final edit.
Crucially, this process is falsifiable and interactive~\citep{qiang2025mle, park2023generative}. If the output does not match the vibe, the Commander provides high-level feedback (``make it darker'', ``increase the tension''), and the agentic system reconfigures the underlying workflow logic rather than just re-rolling a random seed.
{This transition from Stochastic Guessing to Logical Orchestration is what separates Vibe AIGC from current generative tools. We contend that the future of machine learning research lies in perfecting this orchestration layer—enabling AI to not just predict the next token, but to construct the next solution.}

\section{Model-centric Generation}
\label{modelcentric}
\subsection{Prevailing Architectures in Video Generation}

The field of generative AI has progressed from static image synthesis to dynamic video generation, with foundational models now demonstrating significant advancements in text alignment, visual fidelity, motion plausibility, and realism~\citep{liu2025improving}. Current research mainly focuses on text-to-image (T2I)~\citep{rombach2022high}, text-to-video (T2V)~\citep{yin2025towards}, and image-to-video (I2V) generation~\citep{xing2024survey}, as these modalities allow for the collection of scalable, high-quality datasets~\citep{wang2025koala} curated through rigorous filtering and captioning processes~\citep{chen2025avocado}. 

The latest video generation paradigm is the latent diffusion model with a spacetime Transformer, supplanting earlier GAN~\citep{goodfellow2020generative} and VQ-VAE-based~\citep{van2017neural} methods. This approach first compresses a video into a lower-dimensional latent space of ``spacetime patches'' and then employs a diffusion Transformer (DiT)~\citep{peebles2023scalable} to denoise these patches conditioned on a text prompt. This patch-based framework offers the flexibility to process videos of variable resolutions, durations, and aspect ratios within a single model.

For I2V generation, the dominant strategy is to adapt a pre-trained T2V model. For example, Stable Video Diffusion~\citep{blattmann2023stable} fine-tunes the Stable Diffusion model by incorporating temporal layers, enabling it to generate motion that is coherent with a given input frame. Wan compresses the first frame into VAE latents and concatenates them with the noise latent along the channel axis~\citep{wan2025wan}, so that the starting frame would be exactly maintained. A critical challenge in this area is maintaining identity consistency, ensuring a subject's appearance remains stable across frames; techniques like IP-Adapter~\citep{ye2023ip} and integrated memory mechanisms~\citep{yu2025context} have shown promise in addressing this challenge.

However, a fundamental constraint is the immense computational cost of video data, which results in training datasets that are smaller in scale and conceptual breadth than those for LLMs. This creates a disparity between the model's limited world knowledge and high user expectations for both semantic understanding and visual fidelity. Consequently, prompt engineering (PE)~\citep{liu2022design} becomes essential to bridge this gap by aligning user queries with the model's learned data distribution. While sophisticated prompting can unlock impressive reasoning capabilities~\citep{wiedemer2025video}, the generation process often remains stochastic and requires considerable trial-and-error, increasing the barrier to effective use.

\subsection{Editing and Reference-Based Video Generation}

Advancing beyond unconditional generation from a single prompt~\citep{team2025kling}, a significant research frontier is the development of methods for granular control over video generation. Reference-based generation~\citep{ku2024anyv2v} aims to transfer specific attributes from a source to a newly generated video. For instance, style transfer applies the visual aesthetic of a reference image or video onto a target video.
Similarly, subject-driven generation~\citep{team2025klingavatar}, inspired by personalization techniques like DreamBooth~\citep{ruiz2023dreambooth}, first learns a unique identifier for a subject from reference images. 
A diffusion model, conditioned on this identifier and a motion sequence, can then generate a new video of that subject. 
A primary obstacle for all reference-based tasks is the difficulty in acquiring training data; it requires meticulously constructed pairs of reference and target videos with guaranteed correspondence and quality. 
A common failure mode is ``content leakage'', where unintended attributes from the reference are mistakenly rendered in the output. Such artifacts, often stemming from inherent model limitations, are difficult for users to mitigate through prompt engineering alone.

Video editing~\citep{jiang2025vace,wei2025univideo} focuses on modifying an existing video according to user instructions. The most direct approach is text-guided video editing~\citep{he2025openve}, which requires a model to comprehend textual commands. 
Representative methods, such as Senorita~\citep{zi2025se}, focus on object-level editing, including object addition, deletion, or modification. To improve precision, some works utilize masks~\citep{cai2025omnivcus} to define the region for editing, tasking the model with in-painting the masked area conditioned on the text prompt. Similar to reference-based methods, training editing models is constrained by data availability, as paired ``before'' and ``after'' real-world videos hardly exist. This reliance on synthetic training data can lead to artifacts like pixel misalignment and unintended subject alterations.

In practice, many user needs involve a combination of reference-based generation and video editing. Such composite tasks often represent out-of-distribution scenarios for current models, resulting in unpredictable or failed outcomes. Users typically cannot decompose a complex creative intent into a sequence of discrete operations that a model can reliably execute. Exhaustively enumerating all possible task combinations is infeasible, highlighting a fundamental gap between specialized model capabilities and the compositional nature of real-world demand.

\subsection{Unified Architectures for Video Understanding and Generation}

A forward-looking research direction is the development of single, unified models capable of both understanding and generation~\citep{cui2025emu3,lin2025uniworld}. This approach treats video not as a specialized data type but as another modality, akin to text or audio, to be processed within a large-scale, multi-modal architecture~\citep{teng2025magi}.

The mainstream unification tends to be achieved by adopting core architectural principles from LLMs. The critical first step is the discretization of continuous video data into a sequence of tokens. This is typically accomplished using a VQ-VAE or a similar network, which learns a codebook of visual patterns and compresses video patches into discrete codes. Once tokenized, the video sequence, now represented as a series of integers, can be processed by a standard Transformer architecture alongside tokens from other modalities like text. Within this framework, diverse tasks are framed as a unified next-token prediction problem. Consequently, tasks such as T2V, video captioning, video prediction, and video editing can be handled by a single model. However, this approach still faces challenges related to data quality and the need for exhaustive task enumeration. Empirically, these unified models still performs struggling on the fundamental T2V and I2V tasks, lagging behind specialized DiT models in terms of both generation fidelity and semantic alignment~\citep{huang2024vbench,ghosh2023geneval}.


\subsection{Analysis of Realworld Workflows}

Video generation models are increasingly adopted in various application domains, including animation, motion comics, news broadcasting, and film post-production, where they can simplify traditional content creation and reduce costs. However, due to the inherent limitations of current models—such as the stochastic nature of their outputs and unpredictable capabilities—users must devise intricate, multi-step workflows to achieve desired results. 
For instance, a creator producing a short-form drama might first develop a storyboard, generate a keyframe for each shot, and then iteratively refine prompts to produce short video segments, typically 5 or 10 seconds in length~\citep{wan2025wan,kong2024hunyuanvideo}, as constrained by the video generation service.
These segments often suffer from artifacts~\citep{ye2025realgen} like color discrepancies, inconsistent character identity, poor realism, and incorrect pacing or lip synchronization~\citep{hu2025harmony} (even with audio-video joint generation~\citep{huang2025jova}).
Consequently, significant manual post-processing is required to edit and assemble these clips into a coherent whole, followed by audio dubbing and video super-resolution~\citep{du2025unimmvsr}. This process is inefficient, with substantial time and computational resources expended on failed generation attempts, making the outcome heavily reliant on trial and error.
While users can produce a viable prototype, a more intelligent system is needed: one that can comprehend the user's high-level creative intent, automatically decompose the task, and orchestrate the use of various tools, rather than depending solely on the combination of raw model outputs and manual prompt engineering.

\section{Preliminary Attempts}
\label{attempts}
Prior to formalizing the {Vibe AIGC} paradigm, researchers conducted a series of exploratory studies across various domains. These preliminary attempts mark a critical transition from model-centric generation to agentic orchestration.

\subsection{Vibe AI-Generated Text Content}

\paragraph{Deep Research: Agentic Synthesis of Creative Context} The inception of any creative work requires a deep understanding of the underlying subject matter. Traditional AIGC workflows often suffer from ``contextual shallowness'' due to the static nature of pre-trained knowledge. Recent advancements in autonomous reasoning, most notably {OpenAI's Deep Research} ~\citep{openai2025deepresearch}, have demonstrated the potential for long-horizon information synthesis. By employing LLM-based agents to perform multi-step web searches, cross-verify disparate sources, and synthesize comprehensive knowledge bases, these systems build a robust semantic foundation before content generation begins. This ``thinking-before-creating'' paradigm is essential for Vibe AIGC, as it ensures that the generated content is grounded in a sophisticated aesthetic and factual context, moving beyond the limitations of single-model-based generation.

\begin{figure}[!h]
    \centering
    \includegraphics[width=0.9\linewidth]{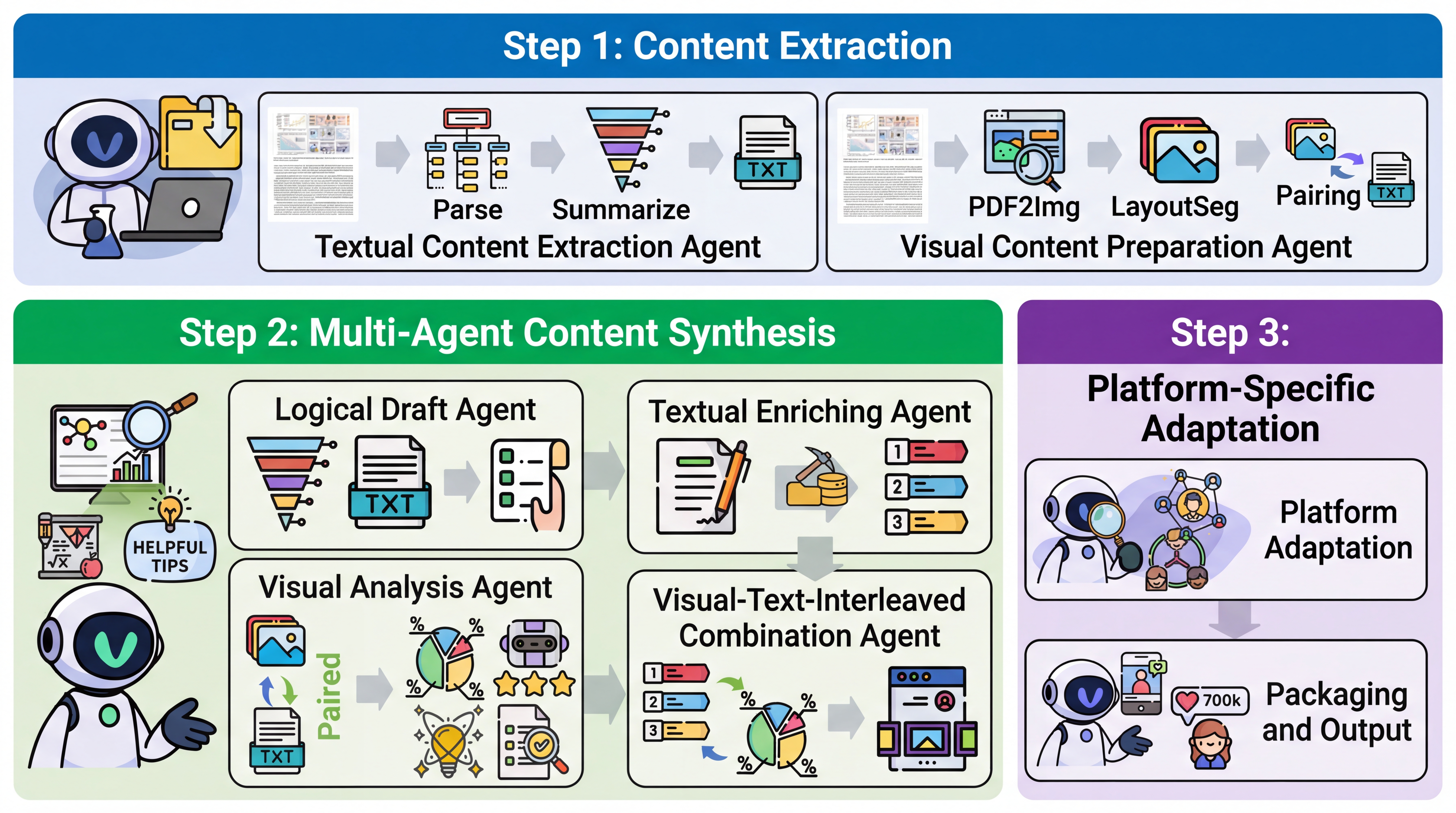}
    \caption{A collaborative multi-agent pipeline in AutoPR.}
    \label{fig:autopr}
\end{figure}

\paragraph{AutoPR: From Fragmented Manual Promotion to One-Click Agentic Pipeline}

In the traditional scholarly promotion workflow, researchers are often reduced to ``manual dispatchers''. Even with the assistance of LLMs, the process remains model-centric and fragmented: authors must manually juggle multiple LLM interfaces for summarization, download and crop figures from PDFs, and painstakingly rewrite content to satisfy the distinct technical constraints and ``vibes'' of various social platforms. 
Thus, researchers introduced \textbf{AutoPR} ~\citep{chen2025autoprletsautomateacademic} in Figure~\ref{fig:autopr}, a novel task that formalizes the transformation of research papers into accurate, engaging public content, and proposed a collaborative multi-agent system, comprising Logical Draft, Visual Analysis, and Textual Enriching agents.

\subsection{Vibe AI-Generated Image Content}

\paragraph{Poster Copilot: Layout Reasoning and Aesthetic Control} 

\begin{figure}[!h]
    \centering
    \includegraphics[width=0.9\linewidth]{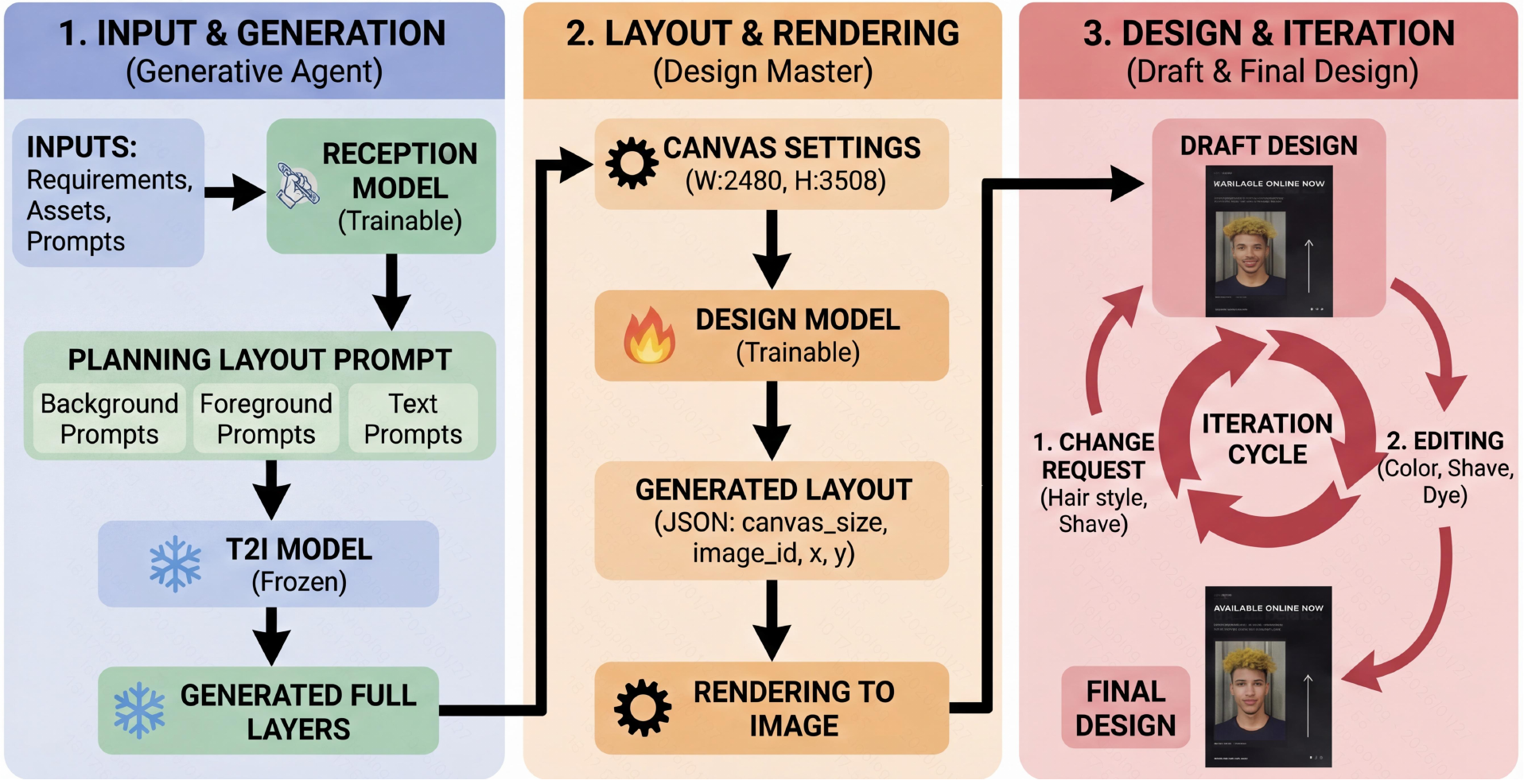}
    \caption{A collaborative multi-agent pipeline in Poster Copilot.}
    \label{fig:postercopilot}
\end{figure}

In professional graphic design, the primary challenge is the precise control over spatial layout and typography. \textbf{Poster Copilot} ~\citep{wei2025postercopilotlayoutreasoningcontrollable} in Figure~\ref{fig:postercopilot} explored agentic layout reasoning and controllable editing. Unlike black-box generators, Poster Copilot functions as a design partner that translates abstract ``Vibe'' instructions into concrete design parameters such as geometric composition, color palettes, and layer hierarchies. By incorporating a feedback loop with human-in-the-loop editing, this system demonstrates how agents can bridge the gap between vague human aesthetic preferences and the rigid technical requirements of professional design.


\subsection{Vibe AI-Generated Video Content}

\paragraph{AutoMV: Multi-Agent Orchestration for Music-to-Video Generation}

\begin{figure}[!h]
    \centering
    \includegraphics[width=0.9\linewidth]{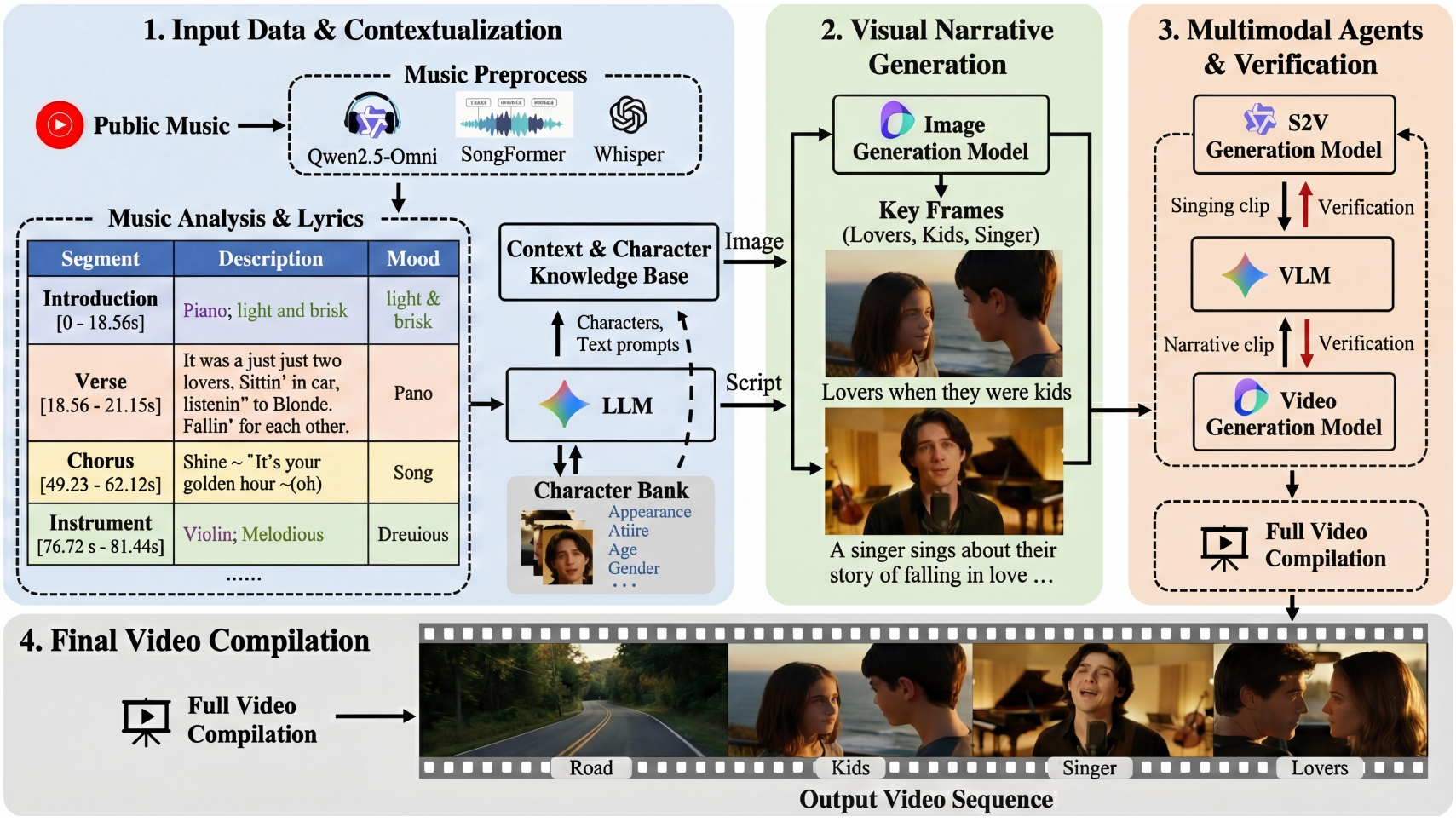}
    \caption{A collaborative multi-agent pipeline in AutoMV.}
    \label{fig:automv}
\end{figure}

To tackle the complexities of music video (MV) creation, where visuals must align with rhythm, lyrics, and emotional arcs, researchers developed \textbf{AutoMV} ~\citep{tang2025automvautomaticmultiagentmusic}. In Figure \ref{fig:automv}. AutoMV represents a shift toward a collaborative multi-agent pipeline. The system employs a Screenwriter Agent to draft narrative scripts based on musical attributes (e.g., beats and structure) and a Director Agent to manage a shared Character Bank and coordinate with various video generation tools. This framework ensures that different segments of a full-length song remain visually and stylistically consistent. The success of AutoMV underscores the necessity of a modular, role-playing agentic structure in managing the high-level intent of a creative project.

\medskip In addition to the aforementioned works, there are numerous studies such as MotivGraph-SoIQ~\citep{lei2025motivgraphsoiqintegratingmotivationalknowledge}, VideoAgent~\citep{wang2024videoagentlongformvideounderstanding}, HollywoodTown~\citep{wei2025hollywoodtownlongvideogeneration}, and LVAS-Agent~\citep{zhang2025longvideoaudiosynthesismultiagent}. All  preliminary efforts reveal a clear trajectory: the future of AIGC lies in orchestrating specialized agents capable of reasoning, planning, and maintaining long-term consistency. However, these systems remain largely fragmented within their respective domains. This observation serves as the primary catalyst for {Vibe AIGC}, which seeks to unify these agentic capabilities into a cohesive, intent-driven ecosystem.

\begin{figure*}[t]
    \centering
    \includegraphics[width=1\linewidth]{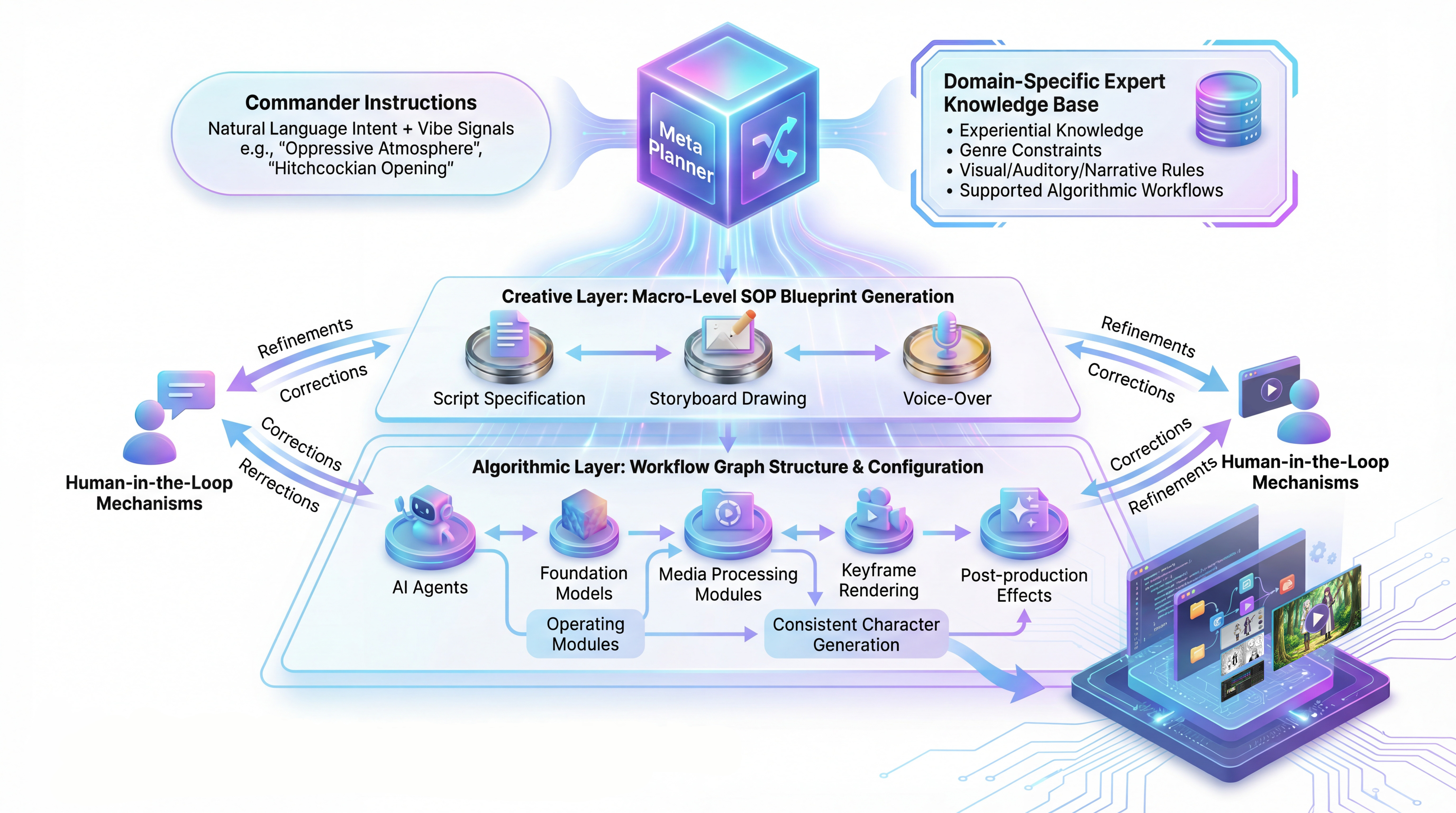}
    \caption{Schematic diagram of Vibe AIGC architecture.}
    \label{fig:vibe_aigc}
\end{figure*}

\section{Vibe AIGC}
\label{vibeaigc}
While the SOP-based fixed patterns and manual orchestration modes aforementioned have, to some extent, mitigated the uncertainty of single-prompting and the rigidity of fixed workflows, they fundamentally remain constrained by a tool-centric bottleneck.
These paradigms necessitate deep technical expertise for tool selection and graph construction, plunging users into a ``cognitive misalignment": they are ensnared by the complexities of low-level technical implementation rather than focusing on core creative expression.

In practice, AIGS content creation is characterized by two features: fine-grained, diversified requirements and a demand for the encapsulation of technical execution. Creators’ intentions are often highly abstract and multifaceted, exceeding the reach of finite, static SOPs; concurrently, users prioritize ``intent fulfillment" over ``tool scheduling." This supply-demand mismatch creates a binary dilemma for existing methods: they either suffer from task failure due to the limited generalization of SOPs or impose excessive overhead due to the high cognitive load of manual orchestration.

Given these challenges, to truly realize the ``User as Commander" vision advocated by Vibe Coding within the AIGC domain, we propose a novel top-level design: Vibe AIGC.
This chapter details the architecture, the core of which is a paradigm shift from ``executing preset processes" to ``autonomously constructing solutions." Under this framework, natural language is no longer merely a prompt but is compiled into meta-instructions for executable workflows. The system evolves from a single-model inference engine or a rigid workflow framework into a self-organizing multi-agent orchestration system driven by a Meta Planner, incorporating human-in-the-loop mechanisms.

\subsection{Top-Level Architecture}

The high-level design of Vibe AIGC aims to establish a system-level semantic entropy reduction mechanism, bridging the gap between unstructured, high-dimensional creative intent and structured, deterministic engineering implementation. In traditional paradigms, this entropy reduction process relies entirely on the user, who must manually translate requirements into tool selection and configuration. As shown in Figure \ref{fig:vibe_aigc}, in contrast, the Vibe AIGC architecture externalizes this cognitive transformation by constructing a closed-loop system centered on the Meta Planner, supported by a Domain-Specific Expert Knowledge Base, and directed toward hierarchical workflow orchestration.

At the apex of this architecture, the Meta Planner serves as the primary commander at the human-computer interface. Rather than executing generation tasks, it is responsible for receiving natural language and translating it into global system scheduling~\citep{jiang2025screencoderadvancingvisualtocodegeneration}. This process transcends simple keyword matching, employing reasoning-based dynamic construction~\citep{xiong2025selforganizingagentnetworkllmbased}. To ensure professional precision, the Meta Planner interacts deeply with an external domain-specific knowledge, storing professional skills, experiential knowledge, and a comprehensive registry of supported algorithmic workflows. For instance, when the Planner receives a request for an ``oppressive atmosphere'', it queries the knowledge base to deconstruct this abstract ``vibe" into specific engineering constraints—such as low-key lighting, close-ups, and low-saturation filters—thereby mitigating hallucinations and mediocrity common in general LLMs~\citep{li2025agentorientedplanningmultiagentsystems}.

Regarding execution logic, the high-level design adopts a hierarchical orchestration strategy, mapping complex generation tasks through top-down layers of abstraction. The Meta Planner first generates a macro-level SOP blueprint at the creative layer, then propagates this logic to the algorithmic layer to automatically derive and configure the workflow graph structure~\citep{qiu2025blueprintfirstmodelsecond}. This hierarchical design ensures the system can both comprehend the ``director’s vision'' at a macro level and precisely control ``technician operations'' at a micro level. In essence, the top-level design of Vibe AIGC is not a static toolkit but a dynamic decision flow driven by the Meta Planner: it perceives the user's ``Vibe" in real-time, disambiguates intent via expert knowledge, and ultimately grows a precise, executable workflow from the top down~\citep{xiong2025selforganizingagentnetworkllmbased}.

\subsection{Meta Planner}

As the cognitive core of the Vibe AIGC architecture, the Meta Planner assumes the critical responsibility of translating natural language intent into an executable system architecture. Departing from the paradigm of traditional Large Language Models (LLMs) that function merely as text generators or simple routers, the Meta Planner is engineered as a ``System Architect" endowed with high-level reasoning capabilities~\citep{goebel2025llmreasoningmodelsreplaceclassical}. Positioned at the forefront of human-computer interaction, it directly interfaces with the user’s natural language input. Its primary function is not the immediate generation of content, but rather deep intent parsing and task decomposition. Upon receiving a user’s ``Commander Instructions"—which are often ambiguous and unstructured—the Planner identifies explicit functional requirements while simultaneously capturing latent "Vibe" signals, such as style, mood, and rhythm. Through multi-hop reasoning, it converts these signals into internal logical representations, thereby triggering the entire generative engine.

\subsection{Intent Understanding}

The reasoning depth of the Meta Planner stems from its tight synergy with the domain-expert knowledge. This process is designed to flesh out sparse, subjective user instructions into actionable and objective creative schemes, thereby addressing intent sparsity~\citep{fagnoni2025opuspromptintentionframework}.

The Planner begins by querying the creative expert knowledge modules, which encapsulate a vast array of multi-disciplinary expertise. For instance, when a user provides the instruction for ``a Hitchcockian suspenseful opening," the Planner does not merely treat it as a text prompt. Instead, it leverages the knowledge base to deconstruct the abstract concept of ``suspense" into a series of precise creative constraints: visually, it mandates "dolly zoom" camera movements and ``high-contrast lighting"; auditorily, it requires "dissonant intervals" in the score; and narratively, it dictates an editing rhythm based on ``information asymmetry." Through this process, the Planner externalizes implicit knowledge, transforming the user’s subjective aesthetic intuition into objective, concrete creative scripts. Consequently, this mitigates the issues of ``averaging" (mediocrity) or "hallucination" often found in traditional AIGC tools due to comprehension gaps~\citep{shi2025flowagentachievingcomplianceflexibility}.

\subsection{Agentic Orchestration}

Upon the completion of intent expansion and disambiguation, the system enters the stage of Agentic Orchestration. The role of the Meta Planner shifts from creative director to dynamic compiler. It constructs the system by mapping the aforementioned creative scripts into specific algorithmic execution paths, based on the algorithmic workflows and tool definitions stored in the knowledge base~\citep{li2024autoflowautomatedworkflowgeneration}.

The Planner traverses the system’s atomic tool library—which includes various Agents, foundation models, and media processing modules—to select the optimal ensemble of components and define their data-flow topology. It possesses adaptive reasoning for task complexity: for simple image generation, it may configure a linear text-to-image pipeline; whereas for long-form video production, it autonomously assembles a complex graph structure encompassing script decomposition, consistent character generation, keyframe rendering, frame interpolation, and post-production effects. Crucially, this orchestration includes the precision configuration of operational hyperparameters (e.g., sampling steps and denoising strength). Ultimately, the Meta Planner generates a complete, logically verified set of executable workflow code, achieving an automated leap from natural language concepts to industrial-grade engineering implementation.

\section{Limitations}
\label{views}

\noindent\textbf{The ``Bitter Lesson'' and Model Centrality.}
The ``Intent-Execution Gap'' is not a permanent architectural flaw, but a temporary symptom of insufficient model scale~\citep{sutton2019bitter}. If a single foundational model eventually achieves a near-perfect internal world model, the need for a complex orchestration layer may vanish. In this view, ``Vibe'' is merely a high-entropy prompt that current models cannot yet parse, but future models will execute in a single shot without the overhead of multi-agent delegation.

\noindent\textbf{The Paradox of Control: Commander vs. Craftsman.}
The shift from prompt engineer to Commander assumes that users prefer high-level intent over granular manipulation. However, professional creators often require ``pixel-perfect'' control that natural language may inherently lack. Critics argue that Vibe AIGC risks introducing a ``Black Box of Intent''.
By abstracting away the ``How'', we may be trading professional precision for amateur convenience, potentially leading to a ``homogenization of aesthetic'' where the AI's interpretation of a vibe overrides the human's unique creative signature~\citep{flusser2013towards,benjamin2018work}.

\noindent\textbf{The Verification Crisis: Binary Success vs. Aesthetic Subjectivity.}
A fundamental challenge for Vibe AIGC is the absence of a deterministic feedback loop. 
In coding, code either compiles and passes unit tests, or it fails; this verification allows LLMs to iteratively converge on a ground-truth solution~\citep{xu2025swecompassunifiedevaluationagentic}. In contrast, a ``Vibe'' in AIGC is inherently subjective and lacks a formal specification. There is no universal unit test for a ``cinematic atmosphere'' or ``melancholic pacing''. 
Without an objective verification oracle, the recursive orchestration layer may drift into ``aesthetic hallucination'' and fail to meet the user’s unstated creative intent.

\noindent\textbf{Compounding Failures and the Missing ``Compiler''.} 
The reliance on ``recursive orchestration''  introduces significant systemic risks related to error compounding.
In Coding, a compiler acts as a hard constraint that intercepts logical errors. However, Vibe AIGC relies on multiple agents where a minor semantic drift in an upstream agent can lead to catastrophic hallucinations across the entire workflow~\citep{cemri2025multi}. Unlike modular software, generative artifacts often suffer from ``content leakage'' or pixel misalignment that current orchestration layers cannot formally ``debug''. Critics maintain that until an equivalent of an ``Aesthetic Compiler'' is developed, multi-agent workflows will remain a fragile compass for digital construction.

\section{Future directions}
\label{action}
\noindent\textbf{For Researchers: Develop Formal Benchmarks for ``Intent Consistency''}
The current reliance on metrics like FID~\citep{Jayasumana2023RethinkingFT}, CLIP, or perplexity is insufficient for the Vibe AIGC era. We call on the academic community to move beyond evaluating pixel fidelity and instead develop benchmarks that measure Agentic Logic Consistency. We need ``Creative Unit Tests'' that evaluate whether a multi-agent system can successfully decompose a complex, ambiguous ``vibe'' into a logically sound and temporally consistent workflow across multiple modalities.

\noindent\textbf{For Industry Leaders: Incentivize Specialized ``Micro-Foundation'' Models}
The pursuit of a single ``God-model'' that handles all creative tasks is architecturally inefficient for professional workflows. Industry leaders and AI labs should pivot toward training and open-sourcing specialized foundation agents. Rather than monolithic LLMs, the community needs high-performing, lightweight agents trained specifically for niche creative tasks—such as a ``Cinematography Agent'' grounded in film theory, or a ``Creative Director Agent'' for workflow synthesis.

\noindent\textbf{For Software Architects: Standardize Agent Interoperability Protocols.}
The success of Vibe AIGC depends on an evolving ecosystem of collaborative agents. We call for the establishment of Open Agentic Interoperability Standards (e.g., an ``AIGC Protocol''). This would allow agents from different developers to share a common ``Character Bank'', ``Global Style State'', and ``Context Memory'' seamlessly. Without standardized communication protocols, agentic workflows will remain fragmented and closed-source.

\noindent\textbf{For the Data Science Community: Curate Intent-to-Workflow Datasets.}
Current datasets are largely composed of static image-text pairs. Realizing the Vibe AIGC era requires a new class of ``Reasoning-in-the-Loop'' datasets. We need data that maps high-level creative intent to the hierarchical reasoning steps and multi-modal sub-tasks required to achieve it. This will enable the training of Meta-Planners that can ``think before creating''.


\section{Conclusion}
The generative AI community stands at a crossroads where scaling laws alone can no longer bridge the gap between human imagination and machine execution. This paper has argued that the path forward lies in a fundamental paradigm shift from Model-Centric Generation to Vibe AIGC, a framework that reconceptualizes content creation as the autonomous synthesis of multi-agent workflows. 
By closing the Intent-Execution Gap and elevating the user to a Commander, Vibe AIGC offers a necessary solution to the usability ceiling.
As we look toward the next generation of AIGC, our research focus must move beyond the internal weights of models and toward the architecture of orchestration, ensuring that the future of the digital economy is built on a foundation of verifiable intent, long-term consistency, and a truly collaborative human-AI creative process.

\bibliographystyle{unsrtnat}
\bibliography{references} 

\end{document}